\documentclass[runningheads]{llncs}

\usepackage[T1]{fontenc}
\usepackage{graphicx}
\usepackage{hyperref}
\usepackage{multirow}
\usepackage{multicol}
\usepackage{algorithm}
\usepackage[noend]{algpseudocode}

\begin{document}

\title{Data Augmentation For Small Object using Fast AutoAugment}

\author{DaeEun Yoon\inst{1,2}\orcidID{0000-0003-2299-191X} \and
Semin Kim\inst{1,2}\orcidID{0000-0003-3746-0863} \and
SangWook Yoo\inst{1,2}\orcidID{0000-0002-5404-4397} \and
Jongha Lee\inst{1,2}\orcidID{0000-0002-1568-6733}}

\authorrunning{D. Yoon et al.}

\institute{AI R\&D Center of Lululab Inc., Seoul, Republic of Korea \and
\email{\{de.yoon, sm.kim, sangwook.yoo, jongha.lee\}@lulu-lab.com}\\
}

\maketitle

\begin{abstract}
In recent years, there has been tremendous progress in object detection performance. However, despite these advances, the detection performance for small objects is significantly inferior to that of large objects. Detecting small objects is one of the most challenging and important problems in computer vision. To improve the detection performance for small objects, we propose an optimal data augmentation method using Fast AutoAugment. Through our proposed method, we can quickly find optimal augmentation policies that can overcome degradation when detecting small objects, and we achieve a 20\% performance improvement on the DOTA dataset.

\end{abstract}

\begin{figure}[ht]
\includegraphics[width=\textwidth]{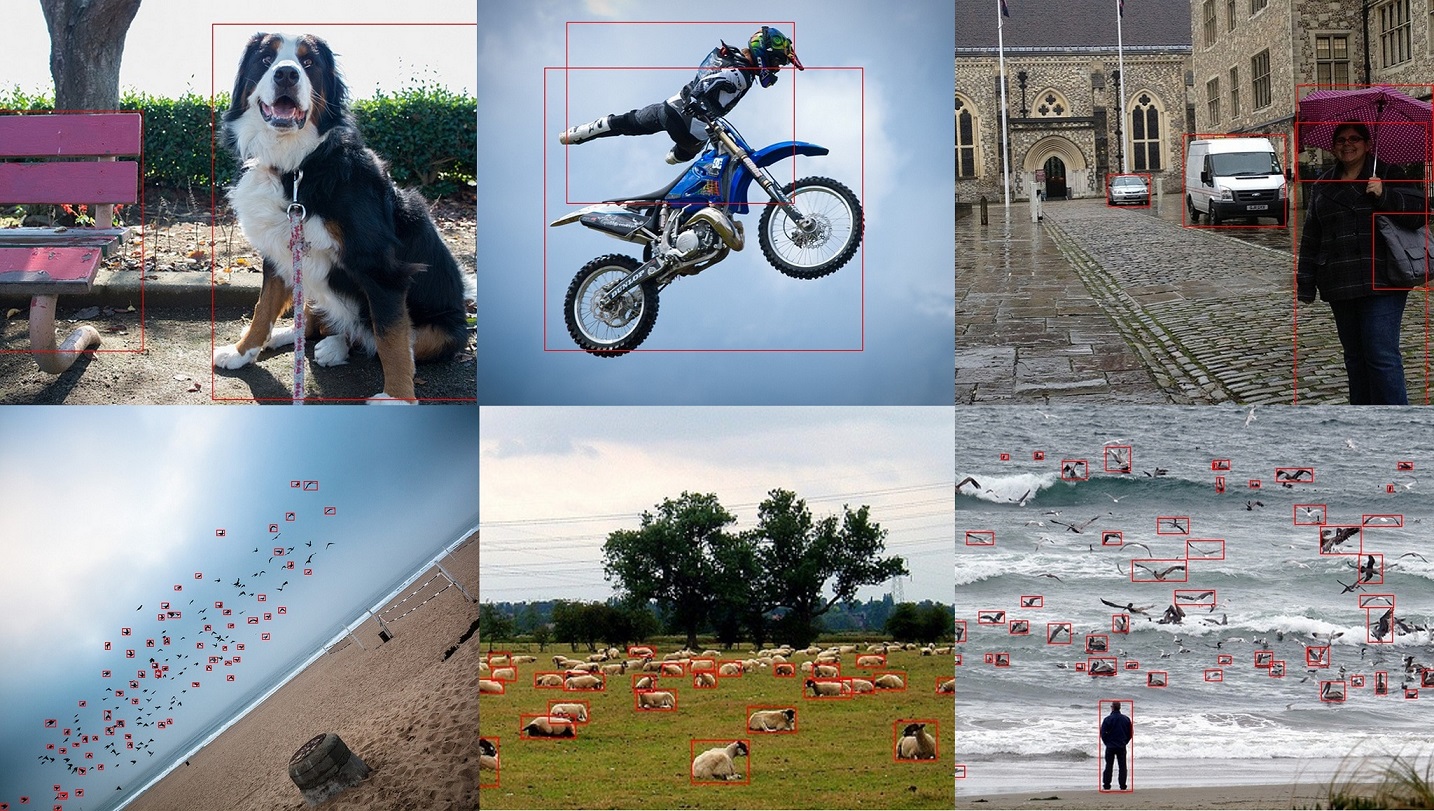}
\caption{An inference sample of Faster R-CNN in MS COCO. The first row is an image consisting of a large object and a medium object instance, and the second row is an image consisting of a small object instance. Despite its clear visibility, small objects have lower detection performance compared to large object detection performance.} 
\label{fig1}
\end{figure}

\section{Introduction}
Through the recent development of deep learning technology, various computer vision tasks have been solved and studied. Among them, object detection is a very important task in computer vision. Object detection has been applied in many areas, including robot vision, autonomous vehicles, satellite image analysis, and medical image analysis, and there have been many advances. However, despite these advances, the problem of detecting small objects has emerged. As shown in Fig. \ref{fig1}, detecting a small object tends to be more difficult than detecting large object or medium object. even in the top submission for the MS COCO \cite{COCO} Object Detection challenge, the performance of detecting small objects is significantly lower than that of detecting large objects. However, detecting small objects is often a more critical task than detecting large objects. For example, if a small forest fire is detected on a real-time mountain CCTV, the spread of a large forest fire can be prevented early, and in the case of self-driving cars, small pedestrians or traffic signs must be detected. Satellite data taken at high altitude should detect small objects, and small defects and objects should be detected in image analysis automation equipment at industrial sites. Moreover, medical images should be able to detect small-sized malignant tumors. Thus, objects that should be detected in the real world are often represented by small pixels in the image. In this paper, there are three perspectives on the degradation of small object detection performance. First, the area of a small object pixel differs from a large object by several times to several tens of times. This data imbalance problem can cause object detection models to be biased towards large objects during training. Second, most data augmentation methods are not effective on small objects. Data augmentation can create models that prevent overfitting and improve generalization performance by adding diverse distributions to training datasets, and consequently contribute significantly to improving the performance of the models. Various augmentation techniques have also been studied in object detection. Pixel-Level transform authorization, which changes pixel values such as RGBSshift, Blur, Random Contrast, and Random Brightness, and geometry transform authorization such as Flip, Shift, and Rotate improve classification or large object detection performance, but not small object detection performance. RandomErase \cite{RandomErase} and Cutout \cite{Cutout} erase or fill parts of the image with specific values, contributing significantly to performance improvement, allowing the model to predict only parts of the image without looking at the entire part of the image, but applying it to small objects is problematic. This is because the operation of erasing a part of the image or filling it with a certain value may be applied to the whole, not to a part of a small object. MixUp \cite{Mixup} improves training performance by blending two images, but does not contribute to improving small object detection performance. CutMix \cite{CutMix} cuts and pastes the image to another image patch. Similarly, it improves overall training performance but does not contribute to improving small object detection performance. The third is the absence of an optimal augmentation policy. Research on the augmentation method of small objects is being conducted steadily. However, most studies do not apply optimal augmentation policies. We propose a novel optimal small object augmentation search method based on the above three perspectives. To evaluate the performance on small object detection, we use the prestigious Fast R-CNN \cite{FasterRCNN} for object detection and perform a quantitative analysis on DOTA \cite{DOTA} Dataset. We have improved the small object detection performance by 20\% compared to before.

\begin{figure}[t]
\includegraphics[width=\textwidth]{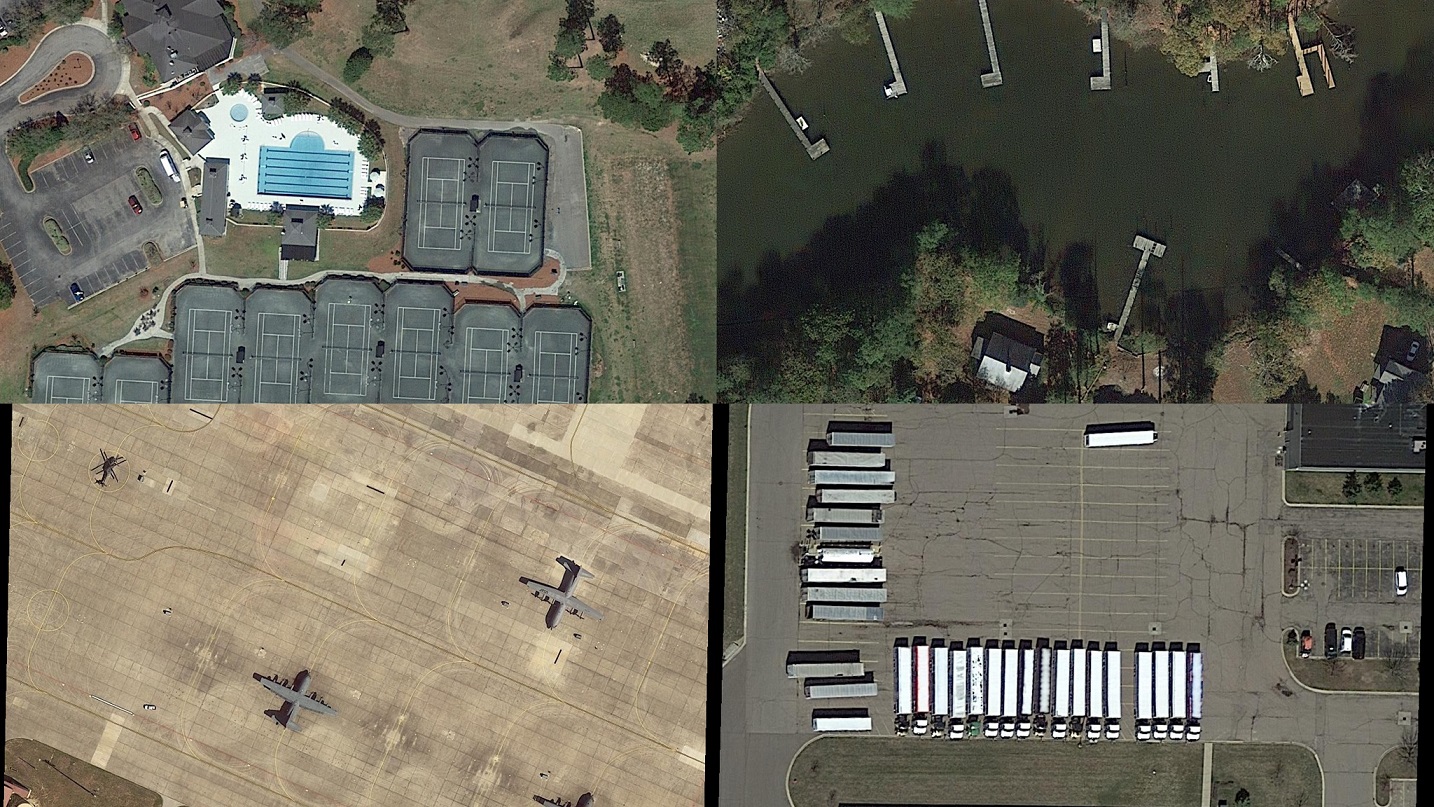}
\caption{Samples from DOTA. It consists of Google Earth, satellite, and aerial images.}
\label{fig2}
\end{figure}

\section{Related Works}
\subsection{Object Detection}
The object detection framework of previous studies is a two-step detector structure, consisting of a region proposal stage that is presumed to have an object and an object classification stage that classifies which category the object is. R-CNN \cite{RCNN} and Fast R-CNN were proposed based on a two-stage detector structure, and later a one-stage detector structure that performs region proposal and object classification at once in a convolution network, representatively YOLO \cite{YOLO}, SSD \cite{SSD}, and RetinaNet. Usually, in a two-stage detector structure, the recognition is performed on an ROI with a specific object, so the accuracy is high but the speed is slow. Conversely, the one-stage detector structure has the advantage of low accuracy but high speed. because region proposal and object classification are performed on the entire image with multiple objects.

\subsection{Small Object Detection}
Several methods have been proposed to improve the performance of Small Object Detection. Scale-Transferrable Object Detection \cite{STDN} proposed a method to generate high-resolution feature maps using the Pixel Shuffler method, which is commonly used in Image Super-Resolution, for small object detection. STDnet \cite{STDnet} proposed a Region Context Network (RCN) that enhances the detection of small objects in high-resolution feature maps. Augmentation for Small Object Detection \cite{SOA} improved small object detection performance by proposing an algorithm that copies and pastes small objects.

\subsection{Small Object Detection in Aerial Images}
The DOTA dataset includes images from Google Earth, GF-2, and aerial(see Fig. \ref{fig2}). DOTA-v2.0 contains 18 common categories, with a total of 11,268 images and 1,793,658 instances. The dataset is divided into four subsets: train, valid, test-dev, and test-challenge. The train subset consists of 1,830 images and 268,627 instances, while the valid subset includes 593 images and 81,048 instances. The test-dev subset has 2,792 images and 353,346 instances, and the test-challenge subset has 6,053 images and 1,090,637 instances. However, ground-truth annotations are not provided for the test-dev and test-challenge subsets.

\begin{table}[t]
\centering
\caption{GPU hours comparison of Fast AutoAugment and AutoAugment, PBA. AutoAugment measured computation cost using an NVIDIA Tesla P100, while PBA measured computation cost using a Titan XP, and Fast AutoAugment estimated computation cost using an NVIDIA Tesla V100.}
\setlength{\tabcolsep}{7pt}
\scalebox{1.0}{
\begin{tabular}[t]{ccccc}
\hline
Dataset & AutoAugment\cite{AutoAugment} & PBA\cite{PBA} & Fast AutoAugment\cite{FastAutoAugment}\\
\hline
CIFAR-10 & 5000 & 5 & 3.5\\
SVHN & 1000 & 1 & 1.5\\
\hline
\end{tabular}
}
\label{table1}
\end{table}%

\subsection{Optimal Augmentation}
Data augmentation has become essential in most machine learning fields. However, determining the appropriate augmentation for dataset is a difficult problem. Although the developer determines the augmentation based on Manual Search or Grid Search, it is not the optimal augmentation suitable for dataset. As a result, active research is being conducted to find the optimal augmentation policy. AutoAugment \cite{AutoAugment} based on reinforcement learning, explored the optimal augmentation policy by giving the child model a test set loss according to the augmentation policy as a reward and achieved state-of-the-art in the classification field. However, this method is time-consuming and costly because the child model must be repeatedly trained to update the policy searching controller (using RNN in AutoAugment). On the other hand, Population Based Augmentation \cite{PBA} is based on the Population Based Training (PBT) algorithm among hyperparameter optimization techniques. Population Based Augmentation(PBA) train several models with different augmentation at the same time, and compare the performance of each model in the middle of training to replicate the parameters of the high-performance model to the parameters of the low-performance model and give some variations of the applied augmentation technique. As shown in Table \ref{table1}, Unlike AutoAugment, time was reduced by 1$/$1000 because repetitive re-training was not required. Also, Fast AutoAugment \cite{FastAutoAugment} uses a trained model without augmentation to obtain an augmentation data loss according to the augmentation policy. It obtains an optimal policy by reducing the density between the original data and the augmented data. Since policy search is conducted using the trained model without repeating re-training, the time is also reduced by 1/1000 compared to AutoAugment.

\section{Method}
In this section, we describe augmentation algorithms for small objects and propose methods and implementations for finding optimal policies.

\subsection{Augmentation Algorithm} \label{augmentationalgorithm}
The augmentation algorithm for small objects is based on copy-pasting strategies used in \cite{SOA}. The algorithm is to copy a small object and paste it to another location. There are three types of methods.

\paragraph{Copy and paste a single object}
Select one small object from the image and paste it to a random location.
\paragraph{Copy and paste multiple objects} 
Select two or more small objects from the image and paste them to a random location.
\paragraph{Copy and paste all objects}
Select all small objects from the image and paste them to a random location. \\

\noindent
The copy-pasting algorithm ensures that the pasted object does not overlap with any existing objects. However, the edge of the copied object may appear unnatural against the background. According to \cite{SOA}, they tested using Gaussian blurring on the edge, but the performance actually declined and the unnatural appearance was still maintained.

\subsection{Searching Policies}
Searching for the optimal augmentation policy is based on Fast AutoAugment. Fast AutoAugment is a method of searching for an augmentation policy that is most suitable for the characteristics of Dataset by estimating density similarity between original data and augmented data. The methodology proposed by Fast AutoAugment for density similarity estimation is that if the augmented data applied with the augmentation policy for the model trained with the original data has a low loss, the optimal augmentation policy. In other words, the lower the loss for the model trained without augmentation, the more similar the density to the original data, and the most appropriate augmentation policy for the characteristics of the dataset.

\subsection{Searching Small Object Augmentation Policies}
SOA \cite{SOA} found the optimal algorithm policy in a way close to Manual search or Grid search by changing the parameter coefficient to improve the performance of small object detection. In this paper, the policy for three copy-pasting algorithms is explored with the Bayesian Optimization TPE \cite{hyperparameter}. As a result, the optimal copy-pasting policy for small object detection is searched.

\begin{figure}[t]
\includegraphics[width=\textwidth]{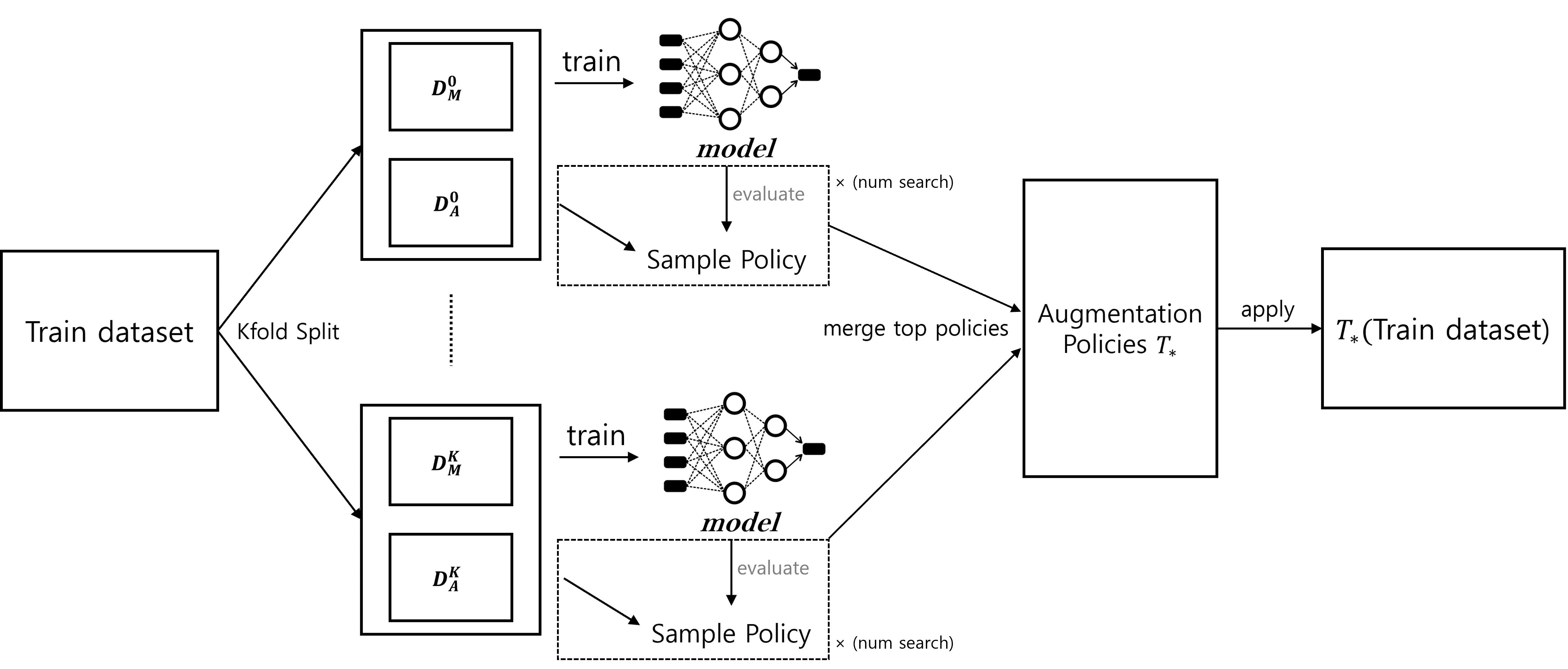}
\caption{An overall procedure of augmentation policy search by Fast AutoAugment algorithm.}
\label{fig3}
\end{figure}

\subsection{Implementation}
Using the Kfold method in the Sklearn \cite{scikit-learn}, the train data is split into $D^{K}_{M}$ and $D^{K}_{A}$(see line 1 in Algorithm \ref{algorithm1}). After that, the model is trained in parallel on each $D^{K}_{M}$ without augmentation (line 3). After training, augmentation policies are searched for using the HyperOpt function in Ray \cite{ray}, which is a library for hyperparameter optimization (line 5-7). The search method is based on the Bayesian optimization TPE \cite{hyperparameter}, and the searching parameters are operation, $p$, and $m$(line 5-7). Operation is the copy-pasting algorithm explained in Subsection \ref{augmentationalgorithm}, $p$ is the copy-pasting probability, and $m$ is the parameter of how many times to paste. In the case of $m$, the parameter is optimized based on the number of times 1 to 3. Then, the searched policy is applied to $D_{A}$ to obtain the loss for the augmentation policy(lines 6-7). For each search, the top N policies with the lowest loss are added to the final policies $T_{*}$ (line 8), and one of the $T_{*}$ policies is randomly chosen and applied as the augmentation policy for each iteration during the final model training. Finally, model is trained by applying the searched policies $T_{*}$(line 9). Fig. \ref{fig3} shows the overall procedure.

\begin{algorithm}[t]
\caption{Implementation pseudo code}
\label{algorithm1}
\begin{algorithmic}[1]
\Require{Train Dataset $D$, $numSearch$, $K$, $N$}
\State Split $D$ into Kfold data $D^{K}_{M}$, $D^{K}_{A}$ 
\For {$k=1,\ldots,K$}
    \State Train Model $M^{k}$ on $D^{k}_{M}$
    \For {$t=0,\ldots,numSearch - 1$}
        \State $T_{t}$ = Search $operation$, $p$, $m$
	\State BayesianOptim($T_{t}$, $Loss(M^{k}|T_{t}(D_{A}))$)
        \State $T^{k}_{t}$ = $T^{k}_{t}$ $\cup$ $T_{t}$
    \EndFor
    \State $T_{*}$ = $T_{t}$ $\cup$ (select top N policies in $T^{k}_{t}$)
 \EndFor
\State Train Model $M$ on $T_{*}(D)$
\end{algorithmic}
\end{algorithm}

\section{Experiments and Results}
\begin{figure}
\includegraphics[width=\textwidth]{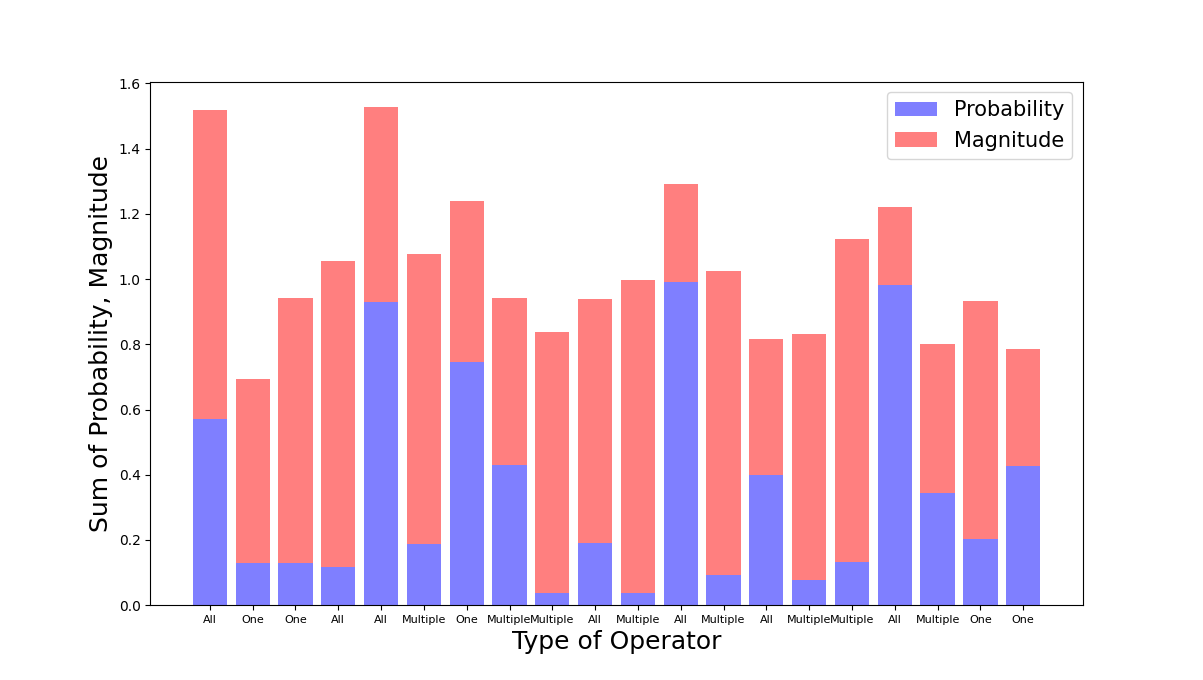}
\caption{The distribution of probability and magnitude of the top 20 policies using Faster R-CNN, where the x-axis represents the type of copy-pasting and the y-axis represents the sum of the parameters p and m. Examining the parameter values, it can be observed that they exhibit an inverse relationship, and that optimal performance is achieved when the values are inversely proportional.}
\label{fig4}
\end{figure}

In this section, we conduct experiments to compare the performance of our proposed methods with the Baseline and SOA, in DOTA-v2.0 valid. Here, baseline is the result of training with the setting and DOTA setting proposed in the object detection model papers, and SOA is the result of applying the best augmentation policy in the SOA paper. The comparison method is Average Precision (AP). The AP scores are calculated separately for four categories: All, Small (object size 0 to 32$\times$32), Medium (object size 32$\times$32 to 96$\times$96), and Large (object size 96$\times$96 or larger). Our proposed method demonstrated a significant improvement in mAP performance, as shown in Tables \ref{table2} and \ref{table3}. Compared to the baseline, our method achieved a 9\% increase in mAP performance in Table \ref{table2}, and a 11\% increase in Table \ref{table3}.  Furthermore, for small objects, our method showed a substantial 20\% improvement in mAP performance in Table \ref{table2}, and a 17\% improvement in Table \ref{table3}. These results highlight the effectiveness of our proposed method in object detection tasks, particularly for detecting small objects. The searched optimal policies are depicted in Fig. \ref{fig4}, showing the distribution of probability($p$) and magnitude($m$). As can be seen from the distribution, probability and magnitude tend to be inversely proportional.

\begin{table}
\centering
\caption{Results of our experiments using Faster R-CNN based on RPN(resnet50 backbone). Experimental results are based on AP(Average Precision) metric.}
\scalebox{1.2}{
\begin{tabular}[t]{lccccc}
\hline
 & mAP & $mAP_{L}$ & $mAP_{M}$ & $mAP_{S}$\\
\hline
baseline \cite{FasterRCNN} & 0.491 & 0.591 & 0.543 & 0.402\\
SOA \cite{SOA} & 0.517 & 0.579 & 0.572 & 0.461\\
Ours & \textbf{0.538} & 0.573 & \textbf{0.578} & \textbf{0.485}\\
\hline
\end{tabular}
}
\label{table2}
\end{table}%

\begin{table}
\centering
\caption{Results of our experiments using RetinaNet(MobileNetV3 backbone). Experimental results are based on AP(Average Precision) metric.}
\scalebox{1.2}{
\begin{tabular}[t]{lccccc}
\hline
 & mAP & $mAP_{L}$ & $mAP_{M}$ & $mAP_{S}$\\
\hline
baseline \cite{RetinaNet} & 0.323 & 0.555 & 0.374 & 0.122\\
SOA \cite{SOA} & 0.346 & 0.544 & 0.433 & 0.132\\
Ours & \textbf{0.359} & \textbf{0.586} & \textbf{0.447} & \textbf{0.143}\\
\hline
\end{tabular}
}
\label{table3}
\end{table}

\section{Conclusion}
We investigated the problem of small object detection. On most datasets, small objects are much smaller than large or intermediate objects, which negatively affected small object detection performance. We introduced a small object augmentation strategy to address this problem and proposed a method to improve small object detection performance by finding an optimal augmentation policy. Our experiments show a mAP 9\% and small object mAP 20\% performance improvement of the proposed method in DOTA-v2.0.

\bibliographystyle{unsrt}
\bibliography{reference}

\begin{thebibliography}{10}

\bibitem{COCO}
Tsung-Yi Lin, Michael Maire, Serge Belongie, James Hays, Pietro Perona, Deva
  Ramanan, Piotr Doll{\'a}r, and C~Lawrence Zitnick.
\newblock Microsoft coco: Common objects in context.
\newblock In {\em European conference on computer vision}, pages 740--755.
  Springer, 2014.

\bibitem{RandomErase}
Zhun Zhong, Liang Zheng, Guoliang Kang, Shaozi Li, and Yi~Yang.
\newblock Random erasing data augmentation.
\newblock In {\em Proceedings of the AAAI conference on artificial
  intelligence}, volume~34, pages 13001--13008, 2020.

\bibitem{Cutout}
Terrance DeVries and Graham~W Taylor.
\newblock Improved regularization of convolutional neural networks with cutout.
\newblock {\em arXiv preprint arXiv:1708.04552}, 2017.

\bibitem{Mixup}
Hongyi Zhang, Moustapha Cisse, Yann~N Dauphin, and David Lopez-Paz.
\newblock mixup: Beyond empirical risk minimization.
\newblock {\em arXiv preprint arXiv:1710.09412}, 2017.

\bibitem{CutMix}
Sangdoo Yun, Dongyoon Han, Seong~Joon Oh, Sanghyuk Chun, Junsuk Choe, and
  Youngjoon Yoo.
\newblock Cutmix: Regularization strategy to train strong classifiers with
  localizable features.
\newblock In {\em Proceedings of the IEEE/CVF international conference on
  computer vision}, pages 6023--6032, 2019.

\bibitem{FasterRCNN}
Shaoqing Ren, Kaiming He, Ross Girshick, and Jian Sun.
\newblock Faster r-cnn: Towards real-time object detection with region proposal
  networks.
\newblock {\em Advances in neural information processing systems}, 28, 2015.

\bibitem{DOTA}
Jian Ding, Nan Xue, Gui-Song Xia, Xiang Bai, Wen Yang, Michael Yang, Serge
  Belongie, Jiebo Luo, Mihai Datcu, Marcello Pelillo, and Liangpei Zhang.
\newblock Object detection in aerial images: A large-scale benchmark and
  challenges.
\newblock {\em IEEE Transactions on Pattern Analysis and Machine Intelligence},
  pages 1--1, 2021.

\bibitem{RCNN}
Ross Girshick, Jeff Donahue, Trevor Darrell, and Jitendra Malik.
\newblock Rich feature hierarchies for accurate object detection and semantic
  segmentation.
\newblock In {\em Proceedings of the IEEE conference on computer vision and
  pattern recognition}, pages 580--587, 2014.

\bibitem{YOLO}
Joseph Redmon, Santosh Divvala, Ross Girshick, and Ali Farhadi.
\newblock You only look once: Unified, real-time object detection.
\newblock In {\em Proceedings of the IEEE conference on computer vision and
  pattern recognition}, pages 779--788, 2016.

\bibitem{SSD}
Wei Liu, Dragomir Anguelov, Dumitru Erhan, Christian Szegedy, Scott Reed,
  Cheng-Yang Fu, and Alexander~C Berg.
\newblock Ssd: Single shot multibox detector.
\newblock In {\em European conference on computer vision}, pages 21--37.
  Springer, 2016.

\bibitem{STDN}
Peng Zhou, Bingbing Ni, Cong Geng, Jianguo Hu, and Yi~Xu.
\newblock Scale-transferrable object detection.
\newblock In {\em proceedings of the IEEE conference on computer vision and
  pattern recognition}, pages 528--537, 2018.

\bibitem{STDnet}
Brais Bosquet, Manuel Mucientes, and V{\'\i}ctor~M Brea.
\newblock Stdnet: A convnet for small target detection.
\newblock In {\em BMVC}, page 253, 2018.

\bibitem{SOA}
Mate Kisantal, Zbigniew Wojna, Jakub Murawski, Jacek Naruniec, and Kyunghyun
  Cho.
\newblock Augmentation for small object detection.
\newblock {\em arXiv preprint arXiv:1902.07296}, 2019.

\bibitem{AutoAugment}
Ekin~D Cubuk, Barret Zoph, Dandelion Mane, Vijay Vasudevan, and Quoc~V Le.
\newblock Autoaugment: Learning augmentation strategies from data.
\newblock In {\em Proceedings of the IEEE/CVF Conference on Computer Vision and
  Pattern Recognition}, pages 113--123, 2019.

\bibitem{PBA}
Daniel Ho, Eric Liang, Xi~Chen, Ion Stoica, and Pieter Abbeel.
\newblock Population based augmentation: Efficient learning of augmentation
  policy schedules.
\newblock In {\em International Conference on Machine Learning}, pages
  2731--2741. PMLR, 2019.

\bibitem{FastAutoAugment}
Sungbin Lim, Ildoo Kim, Taesup Kim, Chiheon Kim, and Sungwoong Kim.
\newblock Fast autoaugment.
\newblock {\em Advances in Neural Information Processing Systems}, 32, 2019.

\bibitem{hyperparameter}
James Bergstra, R{\'e}mi Bardenet, Yoshua Bengio, and Bal{\'a}zs K{\'e}gl.
\newblock Algorithms for hyper-parameter optimization.
\newblock {\em Advances in neural information processing systems}, 24, 2011.

\bibitem{scikit-learn}
F.~Pedregosa, G.~Varoquaux, A.~Gramfort, V.~Michel, B.~Thirion, O.~Grisel,
  M.~Blondel, P.~Prettenhofer, R.~Weiss, V.~Dubourg, J.~Vanderplas, A.~Passos,
  D.~Cournapeau, M.~Brucher, M.~Perrot, and E.~Duchesnay.
\newblock Scikit-learn: Machine learning in {P}ython.
\newblock {\em Journal of Machine Learning Research}, 12:2825--2830, 2011.

\bibitem{ray}
Philipp Moritz, Robert Nishihara, Stephanie Wang, Alexey Tumanov, Richard Liaw,
  Eric Liang, Melih Elibol, Zongheng Yang, William Paul, Michael~I Jordan,
  et~al.
\newblock Ray: A distributed framework for emerging $\{$AI$\}$ applications.
\newblock In {\em 13th USENIX Symposium on Operating Systems Design and
  Implementation (OSDI 18)}, pages 561--577, 2018.

\bibitem{RetinaNet}
Tsung-Yi Lin, Priya Goyal, Ross Girshick, Kaiming He, and Piotr Doll{\'a}r.
\newblock Focal loss for dense object detection.
\newblock In {\em Proceedings of the IEEE international conference on computer
  vision}, pages 2980--2988, 2017.

\end{thebibliography}

\end{document}